\documentclass[runningheads]{llncs} 
\usepackage[T1]{fontenc}

\usepackage{amsmath,amsfonts,bm}

\def\eqref#1{equation~\ref{#1}}

\def\1{\bm{1}}

\DeclareMathAlphabet{\mathsfit}{\encodingdefault}{\sfdefault}{m}{sl}
\SetMathAlphabet{\mathsfit}{bold}{\encodingdefault}{\sfdefault}{bx}{n}

\usepackage{hyperref}
\usepackage{url}
\usepackage{wrapfig}
\usepackage{graphicx}
\usepackage{algorithm}
\usepackage{algpseudocode}
\usepackage{amsmath, bm}
\usepackage{cleveref}
\usepackage{tikz}
\usetikzlibrary{bayesnet}
\usepackage{subcaption}
\usepackage[export]{adjustbox}

\title{Mobile Manipulation with Active Inference for Long-Horizon Rearrangement Tasks}

\titlerunning{Mobile Manipulation with Active Inference}

\author{
Corrado Pezzato\textsuperscript{*} \and
Ozan \c{C}atal\textsuperscript{*} \and
Toon Van de Maele \and
Riddhi J. Pitliya \and
Tim Verbelen
}

\authorrunning{C. Pezzato et al.}

\institute{VERSES AI Research Lab, Los Angeles, California, 90016, USA \\
\email{\{corrado.pezzato,ozan.catal\}@verses.ai} \\
}

\begin{document}

\maketitle
\renewcommand{\thefootnote}{\fnsymbol{footnote}}
\renewcommand{\thefootnote}{\arabic{footnote}}
\begin{abstract}
Despite growing interest in active inference for robotic control, its application to complex, long-horizon tasks remains untested. We address this gap by introducing a fully hierarchical active inference architecture for goal-directed behavior in realistic robotic settings. Our model combines a high-level active inference model that selects among discrete skills realized via a whole-body active inference controller. This unified approach enables flexible skill composition, online adaptability, and recovery from task failures without requiring offline training. Evaluated on the Habitat Benchmark for mobile manipulation, our method outperforms state-of-the-art baselines across the three long-horizon tasks, demonstrating for the first time that active inference can scale to the complexity of modern robotics benchmarks.
\end{abstract}

\section{Introduction}
Active inference offers a principled framework for modeling the action-perception loop, unifying inference and control. Both continuous and discrete formulations have been developed to capture sensorimotor and cognitive processes~\cite{parr2022active,smith2022}, and past works have explored hybrid schemes that integrate discrete decision-making with continuous control in the context of handwriting~\cite {parr2023} and oculomotor tasks~\cite{parr2018active,parr2018discrete}. 

More recently, hybrid continuous-discrete approaches have shown promise in generating rich, goal-directed behavior in 2D simulated settings for reaching, grasping, and tool use~\cite{priorelli2025dynamic,Priorelli2025deep,priorelli2024slow}. However, active inference has yet to demonstrate scalability to the complexity and long time horizons required by modern robotics benchmarks. In particular, no prior work has convincingly shown that active inference alone can match or exceed the performance of state-of-the-art methods in realistic robotic tasks. 

\begin{figure}[t]
    \centering
    \includegraphics[width=\linewidth]{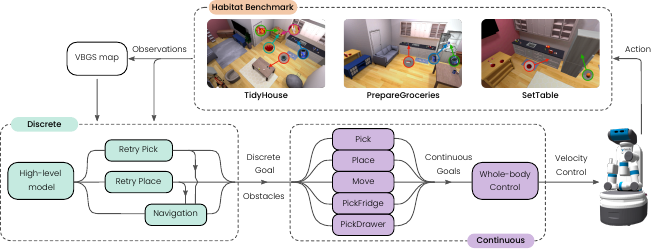}
    \caption{\textbf{Solution overview}. Overview of the proposed solution and the Habitat Tasks described in detail in \cref{sec:methods}.}
    \label{fig:overview}
\end{figure}

In this paper, we address this gap by introducing a fully hierarchical hybrid architecture for active inference-based control in long-horizon mobile manipulation tasks. Fig.~\ref{fig:overview} provides a high-level overview.  Our system combines a high-level active inference agent that reasons over task-relevant abstractions with a novel whole-body controller based on continuous hierarchical active inference~\cite{priorelli_deep_2023,pezzatolearn}. This integration allows for flexible and robust skill execution, supports online adaptation, and eliminates the need for offline training.
We evaluate our approach on three long-horizon mobile manipulation tasks from the Habitat Benchmark~\cite{szot2021habitat}, namely \texttt{TidyHouse}, \texttt{PrepareGroceries}, and \texttt{SetTable}. These tasks require complex, multi-step interactions with articulated objects and constrained environments, such as retrieving items from drawers or refrigerators, transporting them across rooms, and placing them on surfaces. Success demands both long-term planning and precise, reactive motor control. Our method outperforms state-of-the-art baselines across all three tasks, providing a compelling demonstration of active inference.



\subsection{Related work} Long-Horizon mobile manipulation challenges that demand both navigation and manipulation abilities are well-suited to evaluating the effectiveness of embodied AI algorithms. Works like the Habitat Benchmark~\cite{szot2021habitat}, ThreeDWorld~\cite{gan2022threedworld}, and ManipulaTHOR~\cite{ehsani2021manipulathor} require robots to navigate in indoor apartments and rearrange household objects. We chose to focus on the Habitat Benchmark because of its challenging nature: it requires continuous motor control (base and arm), interaction with articulated objects (opening drawers and fridges), and involves complicated scene layouts with clutter. In the literature, long-horizon problems have been tackled with task-and-motion-planning (TAMP) approaches~\cite{kaelbling2013integrated,kaelbling2011hierarchical,srivastava2014combined,garrett2020pddlstream}. 

Although effective, these methods often rely on accurate knowledge of the scene and objects, and are computationally expensive. Learning-based approaches have emerged in recent years as an alternative; however, monolithic end-to-end RL solutions to long-horizon tasks are shown to be prone to failure~\cite{szot2021habitat,gu2022multi}. The main reasons for this are the high sample complexity, inefficient exploration, as well as complicated reward design. 

A common strategy for addressing long-horizon tasks in RL is to decompose them into shorter, more manageable subtasks. For instance, the authors of Habitat~\cite{szot2021habitat} propose a hierarchical framework for mobile manipulation. This integrates classical task planning to generate high-level symbolic goals, while individual low-level skills are trained using RL to achieve these goals. This method demonstrates superior performance compared to monolithic end-to-end RL policies and traditional sense-plan-act pipelines. However, naively chaining multiple skills can result in hand-off issues~\cite{szot2021habitat}, where the terminal state of one skill falls outside the distribution of states encountered during training by the subsequent skill, or leads to states that are infeasible for it to handle. This is especially an issue for stationary manipulation skills. Prior work often decouples the mobile base from the manipulator to simplify the inverse kinematics of redundant systems~\cite{sandakalum2022motion}.

In contrast,~\cite{gu2022multi} highlights that mobile manipulation skills are inherently more robust to error accumulation during skill chaining. By leveraging the robot’s full embodiment, these skills enable more effective subtask execution by allowing the robot to reposition itself. Improved subtask formulation, skill composability, and reusability allowed~\cite{gu2022multi} to reach state-of-the-art performance to date on Habitat.

Active inference offers a distinct perspective on decision-making and control, framing both as aspects of a unified inference process. The theory proposes that complex movements can emerge naturally from generative models that encode goals as prior beliefs and observation preferences~\cite{parr2022active}. Within this framework, the nervous system is seen as maintaining a hierarchical generative model that continuously produces and refines perceptual hypotheses. 

Early work on hybrid active inference combining discrete and continuous processes focused on understanding systems such as oculomotion~\cite{friston2017graphical,parr2018active,parr2018discrete} and handwriting~\cite{parr2023}. For instance,~\cite{friston2017graphical} proposed linking discrete and continuous states by using Bayesian model averaging over discrete priors, and converting the resulting continuous posterior into a discrete representation through Bayesian model comparison. These models typically use a discrete state-space to generate empirical priors, which then guide a continuous controller through sequences of attractive setpoints to achieve articulated behavior. 

Hybrid active inference has been extended to dynamic tasks that demand flexible planning~\cite{priorelli2025dynamic,Priorelli2025deep,priorelli2024slow}. These scenarios require agents to infer object dynamics, decompose tasks into subgoals, and coordinate multiple degrees of freedom to execute composite actions. While such studies demonstrate the promise of hybrid active inference in complex motor control, no implementation has yet scaled to meet the complexity of established robotics benchmarks. In this work, we introduce a fully hierarchical hybrid active inference system that, for the first time, outperforms state-of-the-art baselines on the Habitat Benchmark, demonstrating its viability for long-horizon robotic manipulation.
\subsection{Habitat Benchmark}

The Habitat Benchmark~\cite{szot2021habitat} comprises three long-horizon mobile manipulation tasks visualized in Fig. \ref{fig:overview}: \texttt{TidyHouse}, \texttt{PrepareGroceries}, and \texttt{SetTable}. 

In \texttt{TidyHouse}, the robot is tasked with relocating five objects from an initial to a designated goal position. Both the start and goal locations are typically on open surfaces such as tables or kitchen counters. The \texttt{PrepareGroceries} task involves moving two objects from an already open refrigerator to a countertop and returning one object from the counter back into the fridge. Finally, in \texttt{SetTable}, the agent must move a bowl from a closed drawer to a table, and then place an apple retrieved from a closed fridge on the same table. This scenario involves interacting with articulated objects and picking and placing items within confined containers.
All tasks are specified as a sequence of subgoals. Each subgoal is a tuple $(s_1, s^\ast)$, where $s_1$ is the initial 3D center-of-mass position of an object and $s^\ast$ denotes its goal position. For instance, \texttt{TidyHouse} is defined by a set of five such tuples: $\{(s_1^i, s^{\ast i})\}_{i=1}^5$. The generated scenes for the tasks are built upon the ReplicaCAD dataset, which provides a diverse set of 105 photorealistic indoor environments. Each episode instantiates a rearrangement task by randomly sampling rigid objects from the YCB dataset and placing them on annotated support surfaces, resulting in cluttered initial configurations. 
    
\section{Methods}
\label{sec:methods}
\begin{figure}[t!]
    \centering
    \begin{minipage}[b]{0.49\textwidth}
        \centering
        \includegraphics[width=0.7\textwidth]{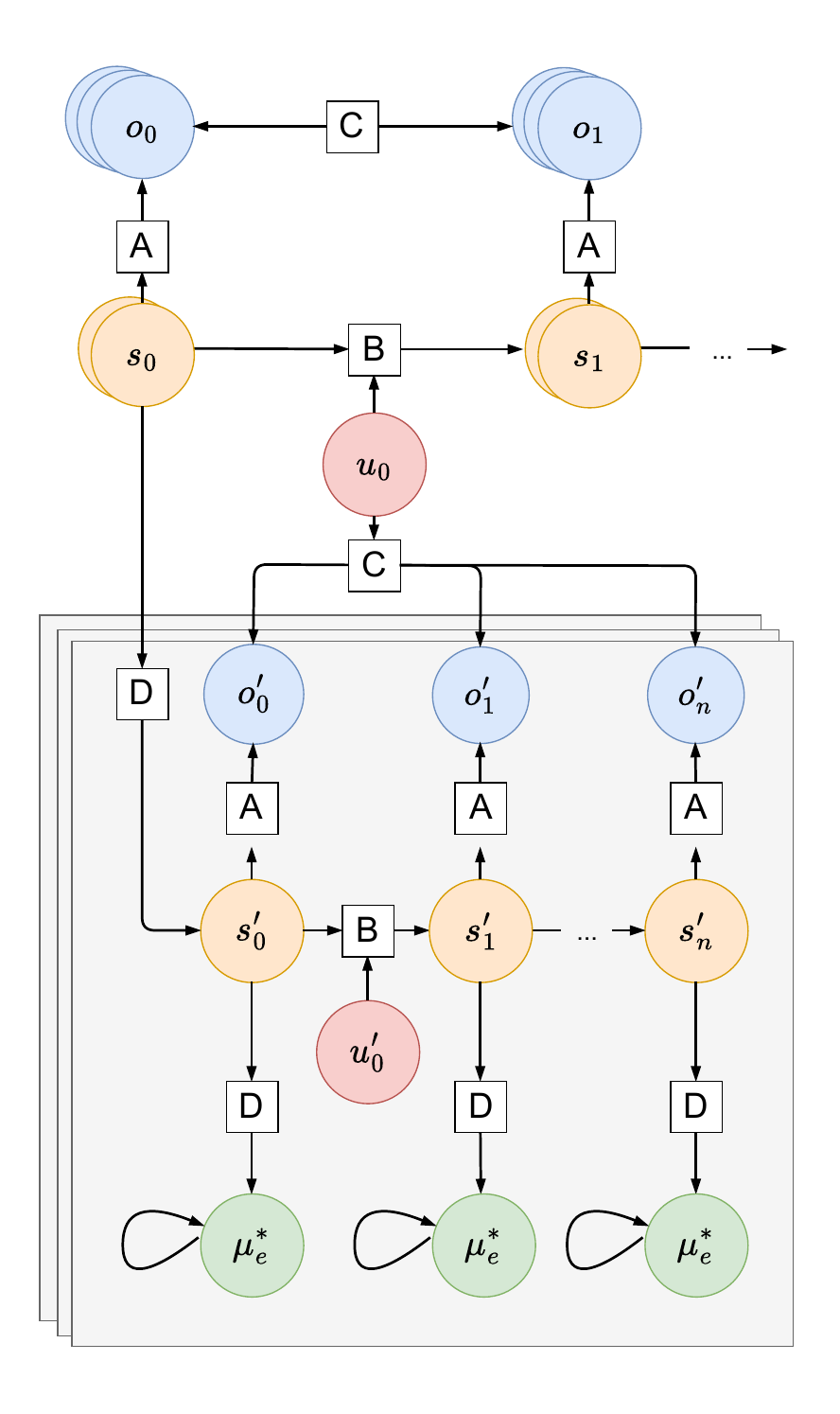}
        \subcaption[]{\label{fig:discrete-hierarchy}}
    \end{minipage}
    \hfill
    \begin{minipage}[b]{0.49\textwidth}
        \centering
        \includegraphics[width=0.9\textwidth]{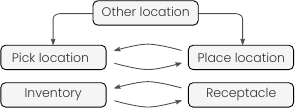}
        \subcaption[]{\label{fig:highlevelplanner}}
        \vspace{2.5em}
        \includegraphics[ width=0.9\textwidth]{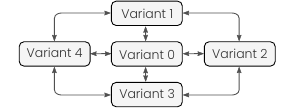}
        \subcaption[]{\label{fig:retryplanner}}
    \end{minipage}
    \caption{\textbf{The Generative Model.} (a) The high-level model sequences skills, each implemented by a generative model interacting with the continuous controller. (b) Dynamics at the highest level. The highest level models the robot location (top) and the object location (bottom). The robot can be either at an other location -- irrelevant to the task -- a pick location, or a place location. The object can be in the robot's inventory or receptacle (location). (c) Dynamics at the pick \& place level. The model switches between various approach parameters based on feedback from the low-level controller to increase robustness against failures.}
    \label{fig:enter-label}
\end{figure}

\subsection{Planning with Universal Generative Models}
To tackle the Habitat Benchmark, we propose an active inference agent, endowed with a hierarchy of generative models, each minimizing the Variational Free Energy. In this universal generative model~\cite{friston2024structure} actions at one level become the preferences of the level below. As we traverse down the hierarchy, each model's time horizon becomes shorter, and the planning becomes more fine-grained. Each component in the hierarchy has its own set of responsibilities. The high-level model orchestrates the sequence of logical actions to take, i.e., where to move and what to pick or place. The \textit{Navigation model} manages the path planning through the environment, while the \textit{Pick \& Place model} coordinates how targets should be approached or released. Finally, at the bottom of the hierarchy lies the active inference whole-body controller, which calculates joint controls for the robot and performs obstacle avoidance. These models keep track of their surroundings using a Variational Bayes approach to Gaussian splatting~\cite{kerbl3Dgaussians,VandeMaele2024vbgs}, which integrates RGBD observations into a probabilistic world map.

\subsubsection{High Level Model}
At the most abstract level, the agent consists of a partially observed Markov Decision Process (POMDP) capturing dynamics over possible states of the object that need to be collected. This allows for the formation of beliefs over these states and robust planning of the agent's various skills. State and action inference in this POMDP is achieved by minimizing variational free energy according to the FEP~\cite{parr2022active}. Crucially, this model tracks the robot's position with respect to the pick/place location as well as the object's relation to the robot, the pick, and the place location. This is then used to schedule either movement or manipulation skills. 
We present this model using  generic active inference terms in Fig.~\ref{fig:discrete-hierarchy}. In contrast to classical active inference models, the actions in our model are abstractions of the skills they represent. The joint controls are generated by the continuous lowest-level model. 

\subsubsection{Pick \& Place Model}
Every interaction with the environment can fail due to unforeseen circumstances or invalid prior beliefs. To accommodate this, the hierarchy maintains a retry model for the failure-prone skills such as picking and placing. As shown in Fig.~\ref{fig:retryplanner}, this model maintains a set of possible approach directions and switches between them based on detected successes or failures of a pick or place. Each approach parameter represents a goal for the lower level controller that can be enacted from that location.

\subsubsection{Navigation Model}
Crucially, still missing from our description of the hierarchical model is a way to move from one point to another. Discrete active inference models are well suited for this as shown in \cite{catal2025}. However, due to the static nature of the environment as well as the prior knowledge about the object location we opted to use A* pathfinding~\cite{hart1968astar} to generate waypoints that act as extrinsic goals for the low level controller.

\subsection{Perception with Variational Bayes Gaussian Splatting}
The models need to infer the structure of the environment to keep track of obstacles and goals in the world. Following the Variational Bayesian approach used throughout the other models, we use a Variational Bayes Gaussian Splat (VBGS)~\cite{VandeMaele2024vbgs} to build a 3D representation of the world from RGBD observations. In this model, the world is represented as a 6D Gaussian mixture over 3D points in space with corresponding 3D color information; the generative model is updated online from observations using Coordinate Ascent Variational Inference (CAVI)~\cite{beal_variational_nodate,bishop2006prml,blei2016vi} without needing a replay buffer or observation queue. As with a normal 3D Gaussian Splat, the model effectively forms a radiance field that captures the room's free and occupied space. This allows for easy obstacle avoidance further down the hierarchy. The parameters of the distributions of component $k$ (i.e. $\mu_{k}$, $\Sigma_{k}$) that generate $\vec{s}$ and $\vec{c}$ are random variables, $z$ is the associated mixture component for a given data point, dependent on the categorical parameters $\pi$. 
    
    
    
    
    
    
    

\subsection{Continuous control with Active Inference}
\begin{figure}[t!]
    \centering
    \includegraphics[width=0.7\linewidth]{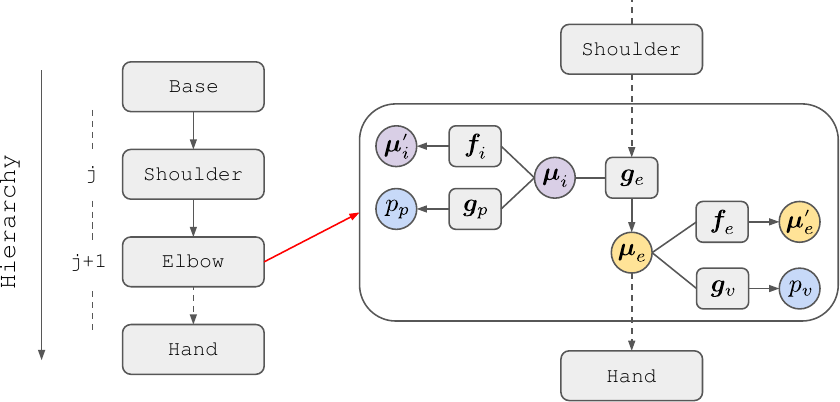}
    \caption{Overview of the Hierarchical Active Inference approach for mobile manipulator control. Intrinsic and extrinsic beliefs $\bm\mu_i$, $\bm\mu_e$ are internal representations of joint angles and Cartesian poses, respectively. They generate proprioceptive and visual predictions $p_p$ and $p_v$ at their level according to the generative models $\bm g_v$, $\bm g_p$. They are also linked through a kinematic generative model $\bm g_e$. The functions $\bm f_i$ and $\bm f_e$ describe the dynamics and are used to guide goal-directed behavior.}
    \label{fig:overview_agent}
\end{figure}

Recent works proposed Hierarchical Active Inference (HAIF) schemes for continuous control of kinematic chains~\cite{priorelli_deep_2023,pezzatolearn}. In this section, we extend 
previous work~\cite{pezzatolearn} to whole-body control of differential drive mobile manipulators. Such robots are composed of a wheeled mobile base and one or more robot arms. We leverage the modularity of HAIF and define one generative model for base control and one for arm control, and then link them through \textit{top-down prediction errors} and \textit{bottom-up predictions}. This results in an overall control scheme that coordinates the whole body of the mobile manipulator at once. An overview of the HAIF approach is depicted in Fig.~\ref{fig:overview_agent}. Importantly, each block in the hierarchy has the same structure and follows the same update rules for state estimation and control. The difference for base and arm control lies in the definition of the generative model $\bm g_e$ and the physical quantities the internal and external beliefs represent, as explained next.
In general, the kinematic generative model $\bm{g}_e$ computes the extrinsic beliefs at the current level~$j$ given the current intrinsic beliefs $\bm\mu_i^{j}$ and the extrinsic beliefs from the level below $\bm\mu_e^{j-1}$, i.e. $\bm\mu_e^{j} = \bm{g}_e(\bm\mu_i^{j}, \bm\mu_e^{j-1})$. The goal is to obtain a set of equations to describe how the internal beliefs of active inference agents are generated and updated over time. Following \cite{priorelli_deep_2023}, the biologically plausible belief update equations are:
\begin{equation}
    \dot{\bm \mu}_i^j = \begin{bmatrix}
        \bm{\mu}_i^{j'} + \pi_p^j\bm\varepsilon_p^j +  \partial_{\bm\mu_i}\bm{g}_e^\top\pi_e^{j+1}\bm\varepsilon_e^{j+1} +  \partial \bm f_i^{j\top}\pi_{\bm\mu_i}^j \bm\varepsilon_{\bm\mu_i}^j\\
        -\pi_{\bm\mu_i}^j\bm\varepsilon_{\bm\mu_i}^j
    \end{bmatrix}
    \label{eq:update_mui}
\end{equation}
\begin{equation}
    \dot{\bm \mu}_e^j = \begin{bmatrix}
        \bm{\mu}_e^{j'} - \pi_e^j\bm\varepsilon_e^j +  \partial_{\bm\mu_e}\bm{g}_e^\top\pi_e^{j+1}\bm\varepsilon_e^{j+1} + \pi_v^{j}\bm\varepsilon_v^{j} +  \partial \bm f_e^{j\top}\pi_{\bm\mu_e}^j\bm\varepsilon_{\bm\mu_e}^j  \\
        -\pi_{\bm\mu_e}^j\bm\varepsilon_{\bm\mu_e}^j
    \end{bmatrix},
    \label{eq:update_mue}
\end{equation}
where $\pi_p,\ \pi_e, \pi_v$ are precision parameters for proprioceptive, extrinsic, and visual models, and $\bm \varepsilon_p ,\ \bm \varepsilon_e$ and $\bm \varepsilon_v$ are the proprioceptive, extrinsic, and visual prediction errors, respectively. Finally, $\bm \varepsilon_{\bm \mu_i}^j = \bm{\mu}_i^{j'} - \bm f_i^j(\bm\mu_i^j)$ and $\bm \varepsilon_{\bm \mu_e}^j = \bm{\mu}_e^{j'} - \bm f_e^j(\bm\mu_i^j)$ are dynamics prediction errors, with precision $\pi_{\bm\mu_i}$ and $\pi_{\bm\mu_e}$. These are used to achieve goal-directed behavior and collision avoidance, as explained later.
In the equations above, we assumed to be able to observe joint positions, velocities, and link positions such that $\bm g_p$ and $\bm g_v$ are identity mappings. Link positions can be computed from joint positions via forward kinematics or estimated via visual input. We now have to find a suitable form for $\bm g_e$ to easily compute the gradients with respect to intrinsic and extrinsic beliefs for both the arm and the base.
\subsubsection{Arm generative model}
Similarly to~\cite{pezzatolearn}, the generative model for the HAIF agent comprises an intrinsic belief $\bm\mu_i$ about joint angles and links' lengths, as well as an extrinsic belief $\bm\mu_e$ about a link's absolute Cartesian position and orientation, for each joint $j$: 
\begin{equation}
    \bm\mu_i^{j} = \begin{bmatrix}
        \theta^{j}, & 
        l^{j}
    \end{bmatrix}^\top \hspace{6mm}
    \bm\mu_e^{j} = \begin{bmatrix}
        x^{j}, & y^{j}, & z^{j}, & q_w^{j}, & q_x^{j}, & q_y^{j}, & q_z^{j}
    \end{bmatrix}^\top = \begin{bmatrix}
        \bm t^j & \bm q^{j}
    \end{bmatrix}^\top.
\end{equation}
The function $\bm{g}_e$ describes the 3D position and orientation of the subsequent link of a kinematic chain given the pose of the previous one. We can define the generative model $\bm g_e$ as in~\cite{pezzatolearn} (see \ref{app:arm_model} for more details): 
\begin{equation}
    \bm{g}_e(\bm\mu_i^{j}, \bm\mu_e^{j-1}) = \begin{bmatrix}
        \bm t^{j-1} + \bm h(\bm q^{j-1}\cdot[0\ \bm t^j]\cdot\bm q^{j-1*}) \\
        \bm q^{j-1}\cdot \bm q^{j}
    \end{bmatrix},
    \label{eq:Tquat}
\end{equation}
where $\bm q^{j-1*}$ is the conjugate quaternion, $''\cdot''$ represents the Hamilton product, and $\bm h(\cdot)$ is a function that returns the imaginary coefficients of a quaternion. In our HAIF agent, the translation $\bm t^{j-1}$ and quaternion $\bm q^{j-1}$ are given by the extrinsic beliefs $\bm \mu_e^{j-1}$. The translation vector~$\bm t^{j}$ and rotation $\bm q^{j}$ are instead dependent on the kinematic properties of the current link $j$ and the joint angle and length $\theta^j,\ l^j$. The generative model can then be fully specified as a function of the intrinsic and extrinsic beliefs, and the gradients can be computed in closed:
\begin{equation}
    \frac{\partial\bm g_e}{\partial\bm \mu_i} = \begin{bmatrix}
        \partial_\theta\bm g_e \\ \partial_l\bm g_e
    \end{bmatrix} \in \mathbb{R}^{2\times7}, \hspace{5 mm} \frac{\partial\bm g_e}{\partial\bm \mu_e} = \begin{bmatrix}
        \partial_{x,y,z}\bm g_e \\ \partial_{\bm q}\bm g_e
    \end{bmatrix} \in \mathbb{R}^{7\times7}
\end{equation}
Thanks to the choice of using quaternions as singularity-free orientation representation, these gradients are easy to compute since the terms in the generative model are either linear or quadratic in the parameters, or they appear as arguments of sine and cosine functions.

\subsubsection{Differential drive generative model}
We now extend the HAIF for robot arm control to a differential drive robot. To do so, we write a simple kinematic model based on Euler's updates where the robot base position and orientation with respect to a world frame are expressed as:
\begin{equation}
    \begin{cases}
        x_{t+1} = x_t + V_t \cos(\theta_t) \delta t\\
        y_{t+1} = y_t + V_t \sin(\theta_t) \delta t\\
        \theta_{t+ 1} = \theta_t + \omega_t\delta t,
    \end{cases}
    \label{eq:kin_model_diff_drive}
\end{equation}
where $V,\ \omega$ are respectively forward and rotational velocities. By considering small wheel increments $\Delta\phi_{R,L} = \phi_{\{R, L\}_t} - \phi_{\{R,L\}_{t-1}}$ in between timesteps where $\phi_{\{R,L\}}$ are the right and left wheel rotations, the expressions for forward and angular velocities result:
\begin{equation}
    V_t = \frac{r}{2\delta t}\begin{pmatrix}
        \Delta\phi_R + \Delta\phi_L
    \end{pmatrix}
    \label{eq:fw_vel}
\end{equation}
\begin{equation}
    \omega_t = \frac{r}{l\delta t}\begin{pmatrix}
        \Delta\phi_R - \Delta\phi_L
    \end{pmatrix}
    \label{eq:fw_rot_vel}
\end{equation}
The terms $r,\ l$ are the wheel radius and distance respectively. The generative model for the differential drive HAIF is defined as a one-level hierarchical model where there are two controllable states, the wheel rotations: 
\begin{equation}
    \bm{g}_e(\bm\mu_i^{j}, \bm\mu_e^{j-1}) = \begin{bmatrix}
        x_{t-1} + \frac{r}{2}\begin{pmatrix}
        \Delta\phi_R + \Delta\phi_L
    \end{pmatrix}\cos(\theta_{t-1})\\
        y_{t-1} + \frac{r}{2}\begin{pmatrix}
        \Delta\phi_R + \Delta\phi_L
    \end{pmatrix}\sin(\theta_{t-1})\\
        \theta_{t-1} + \frac{r}{l}\begin{pmatrix}
        \Delta\phi_R - \Delta\phi_L
        \end{pmatrix}
    \end{bmatrix},
    \label{eq:gen_model_diff_drive}
\end{equation}
We set the intrinsic beliefs to be wheel rotations and extrinsic beliefs to be position $x-y$ and orientation $\theta$ with respect to a world frame:
\begin{equation}
    \bm\mu_i = \begin{bmatrix} \phi_R,\ \phi_L \end{bmatrix}^\top \hspace{6mm}
    \bm\mu_e = \begin{bmatrix}
        x, & y, & \theta
    \end{bmatrix}^\top.
    \label{eq:update_rules_diff_drive}
\end{equation}
The internal and external beliefs are then updated through the gradient of the generative model:
%
\begin{equation}
    \frac{\partial\bm g_e}{\partial\bm \mu_i} = \begin{bmatrix}
        \partial_{\phi_R}\bm g_e \\ \partial_{\phi_L}\bm g_e
    \end{bmatrix} \in \mathbb{R}^{2\times3}, \hspace{5 mm} 
    \frac{\partial\bm g_e}{\partial\bm \mu_e} =\begin{bmatrix}
        \partial_{x}\bm g_e \\ \partial_{y} \bm g_e \\ \partial_{\theta}\bm g_e
    \end{bmatrix} \in \mathbb{R}^{3\times3}
\end{equation}

\subsubsection{Whole-body generative model}
The arm and base models can be combined into a single whole-body model by defining an overall hierarchical structure that combines the two. From \cref{eq:update_mui,eq:update_mue}, one can notice how the prediction errors at the level above $\bm\varepsilon_e^{j+1}$ influence the beliefs at the current level $\dot{\bm \mu}_i^j$ and $\dot{\bm \mu}_e^j$. By this logic, the extrinsic prediction errors of the first level of the hierarchy will not have any influence since there is no level left below in the chain. However, we can propagate these errors back to the top level of the hierarchy of the mobile base kinematic model. By doing so, the base can further minimize free energy by moving its wheels. In turn, we can propagate up from the base kinematic model the predictions about the first link's position and orientation of the robot arm, closing the loop. Mathematically, all equations remain the same apart from the one corresponding to the update of internal beliefs $\dot{\bm \mu}_i^j$ for the base, which becomes:

\begin{equation}
    \dot{\bm \mu}_i^0 = \begin{bmatrix}
        \bm{\mu}_i^{0'} + \pi_p^0\bm\varepsilon_p^0 +  \partial_{\bm\mu_i}\bm{g}_e^\top\pi_e^{1}(\kappa_{base}\bm\varepsilon_{e, base}^{1} + \kappa_{arm}\bm\varepsilon_{e, arm}^{2}) +  \partial \bm f_i^{0\top}\pi_{\bm\mu_i}^0 \bm\varepsilon_{\bm\mu_i}^0\\
        -\pi_{\bm\mu_i}^0\bm\varepsilon_{\bm\mu_i}^0
    \end{bmatrix}
    \label{eq:update_mui_whole_body}
\end{equation}
where $\kappa_{base}$ and $\kappa_{arm}$ are tuning parameters to weight the effect of arm and base prediction errors. By tuning these parameters, one can shape robot behavior, for instance, to allow more or less base response due to the arm's extrinsic prediction errors. The equation for $\dot{\bm \mu}_e^0$ remains the same since the gradient with respect to extrinsic beliefs is zero. This is because the values $x_{t-1}, \ y_{t-1}, \ \theta_{t-1}$ are simply constants from the previous time step and not beliefs from the level below.  

This simple change allows using the base motion to minimize the arm's prediction errors, extending the arm's reachability beyond its stationary workspace. This is crucial for the successful completion of the Habitat Benchmark. Additionally, we still preserve the ability to send individual goals to the base as explained below.

\subsubsection{Goals, obstacles, and control}
To realize goal-directed behavior, we can define attractive goals and repulsive forces as in~\cite{priorelli_deep_2023,pezzatolearn}. Goals can be both intrinsic (joint positions) or extrinsic (Cartesian poses) for the arm and base, and they can be combined to define future desired states $\bm \mu^*$. 
Goals act as attractors, forming dynamic functions  $\bm f_a = \kappa_a(\bm \mu^* - \bm \mu)$ that linearly minimize the distance between the desired and current states. The desired states can be defined flexibly in terms of the current beliefs as
\begin{equation}
    \bm \mu^* = N\bm \mu + \bm n^*,
\end{equation}
where $N$ achieves dynamic behaviors, such as keeping a limb vertical by imposing the $x,\ y$ coordinates of a link to be the same as the previous one, while $ \bm n^*$ imposes an attractor to a static configuration. Some examples of basic goals that can be given to the mobile manipulator in the Habitat Benchmark are 1)~\textbf{End-effector} goal: the robot will use its whole body to reach a target $(x^*,\ y^*,\ z^*)$ position, 2)~\textbf{Base goal}: the robot will move its base to reach a goal $(x^*,\ y^*,\ \theta^*)$, where $\theta^*$ can be updated over time such that the robot faces the goal $\theta^*_t = \arctan2(y^* - y_{t}, x^* - x_{t})$, 3)~\textbf{Arm joint goal}: the robot will reach a specific joint configuration, 4)~\textbf{Combinations of the above}: the user can mix goals for example making the base move while keeping the arm in a certain joint configuration. The same idea of attractive forces can be used for collision avoidance through repulsive forces, where a repulsive state $\bm \mu^!$ has to be avoided (see~\cref{app:collision_avoidance}).  

Given a goal, the control action is computed by minimizing the proprioceptive component of the free energy with respect to the control signals \cite{priorelli_deep_2023}:
\begin{equation}
    \dot{\bm a} = -\partial_a \mathcal{F}_p = -\partial_a\Tilde{\bm s}_p\pi_p\Tilde{\bm\varepsilon}_p
\end{equation}
where $-\partial_a\Tilde{\bm s}_p$ is the partial derivative of proprioceptive observations with respect to the control, and $\bm{\Tilde{\varepsilon}}_p = \Tilde{\bm s}_p - \bm g_p(\bm\mu)$ are the generalized proprioceptive prediction errors.

\subsection{Whole-body high-level skills for objects rearrangement}
Similarly to the chosen Habitat Baseline~\cite{gu2022multi}, we define a set of abstract high-level skills that the high-level planner can sequence at runtime. These skills are implemented with the whole-body controller, and are divided into \texttt{Pick}, \texttt{Place}, \texttt{Move}, \texttt{PickFromFridge},  \texttt{PickFromDrawer}. The skills are defined as a fixed sequence of goals given to the whole-body controller to realize an overall behavior. Every skill computes a whole-body control action for the robot. Details about the skills can be found in~\cref{app:skills}.


\section{Experiments}
\subsection{Experiments setup}

\textbf{Agent embodiment}: The Habitat Benchmark employs a Fetch robot that features a mobile wheeled base, a 7DoF robotic arm, and a parallel-jaw gripper. It is equipped with two RGB-D cameras with a resolution of $128 \times 128$ pixels on both the arm and the head. The robot perceives its state through proprioceptive sensing, which includes joint angles of the arm and the Cartesian coordinates of the end-effector. The robot can also sense the goal positions (3DoF), as well as a scalar to indicate whether an object is held. Obstacle positions are sensed by querying the map model.  

\noindent\textbf{Action space}: The action space is a 10DoF continuous space, for whole-body control. It is composed of forward and angular base velocities, a 7DoF arm velocity action, and a 1DoF gripper action. Grasping is abstract as in previous work~\cite{szot2021habitat,gu2022multi}, such that if the gripper action is positive, the object closest to the end-effector within 15cm will be snapped to the gripper. An object is instead released by a negative action.

\noindent\textbf{Evaluation metrics} Each task in the Habitat Benchmark is composed of a sequence of subtasks that must be completed. As in previous work~\cite{szot2021habitat,gu2022multi}, we measure performance by reporting the completion rate at each subtask stage, with the success rate of the final subtask representing the overall task success. Notably, if the previous subtask has failed, the current subtask is also considered a failure independently of its outcome. At the start of each evaluation episode, the robot's base is placed at a random position and orientation, ensuring no collisions, while the arm begins in a resting configuration.

\noindent\textbf{Baselines} We compare our model against methods in~\cite{gu2022multi}, namely a Monolithic RL approach and a Multi-skill RL Mobile manipulation (MM). The latter had the best performance among several learning-based and classical task and motion planning approaches. The Monolithic RL is an end-to-end RL policy for each complete task (\texttt{TidyHouse}, \texttt{PrepareGroceries}, and \texttt{SetTable}). Different reward functions are selected according to oracle knowledge about the current subtask being executed, such as picking or placing, to train a single policy for such long-horizon tasks.  The Multi-skill RL Mobile manipulation approach, instead, trains different mobile manipulation policies that are then chained by an oracle task planner executed in open loop. For details about the baselines, we refer the reader to~\cite{gu2022multi}. 

\subsection{Results}
\begin{figure}[h]
    \centering
    \includegraphics[width=\linewidth]{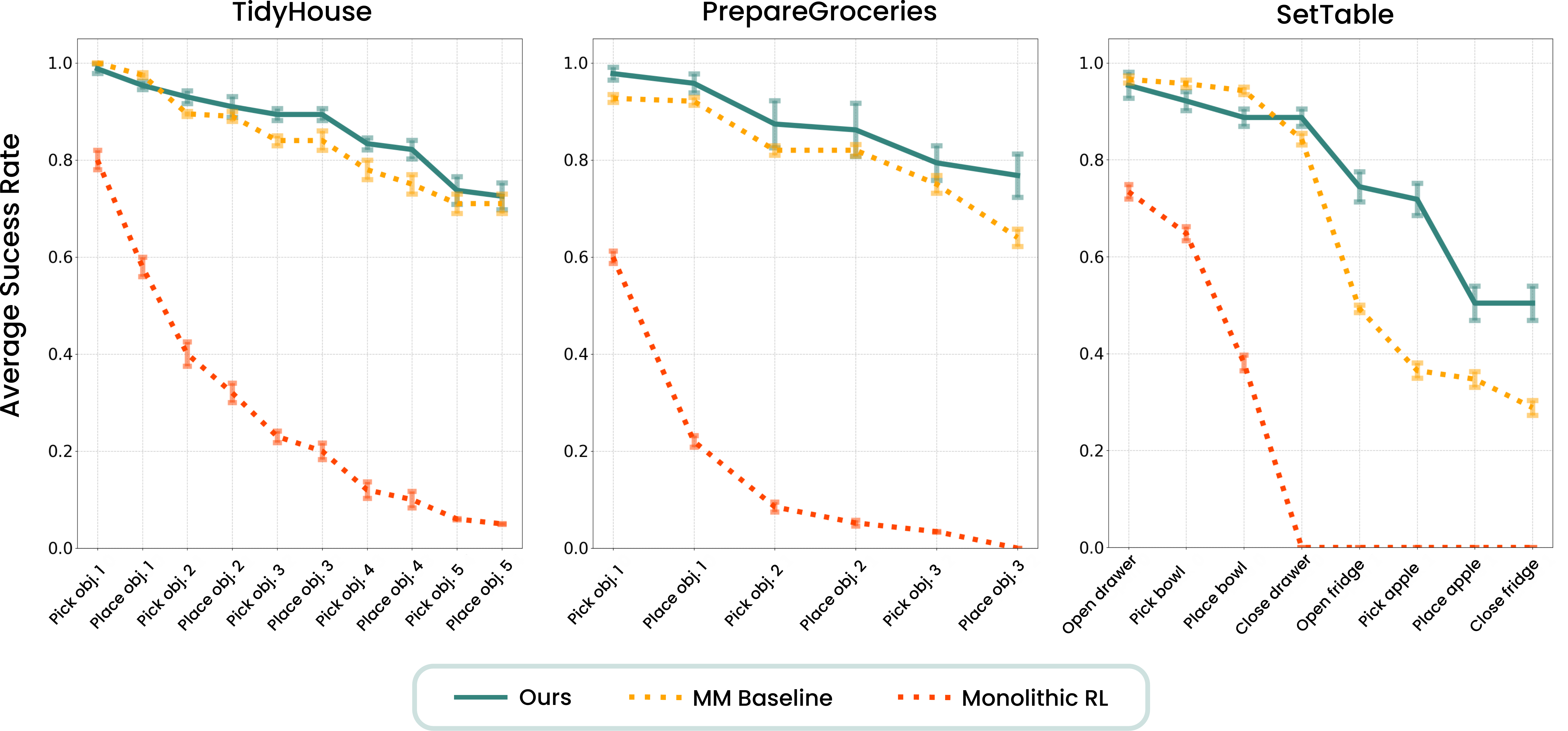}
    \caption{Evaluation results on the Habitat Benchmark, averaged over 100 episodes. Each task is evaluated on different apartment layouts and divided into stages. For a stage to be successful, all previous stages must also be successful. \textbf{TidyHouse}: Evaluated over five pick-and-place stages, from \texttt{Pick obj. 1} to \texttt{Place obj. 5}. \textbf{PrepareGroceries}: Measured from \texttt{Pick obj. 1} to \texttt{Place obj. 3}. \textbf{SetTable}: Involves a more complex sequence including \texttt{Open drawer} $\rightarrow$ \texttt{Pick bowl} $\rightarrow$ \texttt{Place bowl} $\rightarrow$ \texttt{Close drawer}, and similarly for the fridge and apple.}
    \label{fig:results}
\end{figure}

We report the benchmark results in Fig.~\ref{fig:results}. Our approach outperforms the baselines in all three tasks, averaging 72.5\% completion rate in \texttt{TidyHouse}, 77\% completion rate in \texttt{PrepareGroceries}, and 50\% completion rate in \texttt{SetTable} over five seeds. The best performing baseline, MM, instead, averages 71\%, 64\%, and 29\% respectively. Considering all three tasks combined, we achieved a 66.5\% success rate compared to the 54.7\% of the MM baseline. Notably, the MM baseline requires extensive offline training. That is 6400 episodes per task across varied layouts and configurations in the Habitat environments, and 100 million steps per skill across a total of 7 skills. In contrast, our method relies on hand-tuning each skill over just a handful of episodes and is evaluated directly on unseen layouts and configurations, demonstrating strong generalization without the need for data-intensive training. However, we still rely on privileged information, such as the floor map for path planning and articulated object states. These assumptions will be removed in future work. 

\section{Discussion and Conclusion}
In this work, we proposed a hierarchical active inference model to address the Habitat Benchmark, surpassing state-of-the-art performance across all three benchmark tasks. Our system is composed of two key components: a high-level model that selects appropriate low-level skills based on discrete observations, and a set of low-level skills defined through goals for a novel whole-body controller using hierarchical active inference. This architecture enables the system to flexibly adapt to task failures and dynamically adjust behavior in response to environmental changes. Importantly, our method operates online, without requiring offline training, and supports real-time adaptation of the high-level plan. While our current implementation still relies on certain privileged information (such as access to a global map for path planning), our future work will focus on enabling the agent to actively explore and construct maps in real-time. Additionally, knowledge of the state of articulated objects, such as drawers and refrigerators, will be inferred directly from raw RGBD sensory input. Moreover, while low-level skills are currently composed of a fixed sequence of goals for the continuous whole-body controller, one could add an intermediate hierarchical level as in~\cite{Priorelli2025deep} to smoothly transition between subgoals. Another interesting direction to explore would be learning skills directly from demonstration. In summary, our hierarchical hybrid active inference model demonstrates promising results in goal-directed robotic control within complex environments. With further enhancements in perception, exploration, and skill acquisition, we believe this framework could serve as a foundation for more generalized and scalable robotic agents.




\bibliography{mybib}
\bibliographystyle{splncs04}

\appendix

\newpage
\section{Appendix}

\subsection{Kinematic generative model}
To generative model in~\cref{eq:Tquat} is defined from generic transformation matrices. To compute the absolute position and orientation of the current link $j$ given the absolute position and orientation of the previous one, we can write:
\begin{equation}
    ^{w}T_j = \left[
    \begin{array}{c|c}
        R^{j-1}R^j & \bm t^{j-1} + R^{j-1}\bm t^j \\
        \hline
        \bm 0 & 1
    \end{array}
    \right],
    \label{eq:Tgeneral}
\end{equation}
where $w$ indicates the world frame as an absolute reference, $R$ represents a rotation matrix, and $\bm t$ a translation vector. The world frame can be the base link of a robot arm. From \cref{eq:Tgeneral}, we note that the resulting absolute rotation of a link is the multiplication of two rotation matrices. However, we can express this as a quaternion multiplication $\bm q^{j-1}\cdot \bm q^{j}$. Similarly, we can rotate a vector $\bm t^j$ by a quaternion $\bm q^{j-1}$ corresponding to $R^{j-1}$, leading to~\cref{eq:Tquat}. Considering a generic Denavit–Hartenberg (DH) transformation matrix
\begin{equation}
\left[
\begin{array}{ccc|c}
\cos\theta^j & -\sin\theta^j \cos\alpha^j & \sin\theta^j \sin\alpha^j & l^j \cos\theta^j \\
\sin\theta^j & \cos\theta^j \cos\alpha^j & -\cos\theta^j \sin\alpha^j & l^j \sin\theta^j \\
0 & \sin\alpha^j & \cos\alpha^j & d^j \\
\hline
0 & 0 & 0 & 1
\end{array}
\right],
\label{eq:dh}
\end{equation}
we note that the translation vector is simply $\bm t^{j} = [l^j \cos\theta^j, l^j \sin\theta^j, d^j]$. According to the DH convention, the rotational part of the transformation matrix is the composition of a rotation $\theta^j$ about the previous $z$-axis and a rotation of $\alpha^j$ around the $x$-axis. We can then write:
\begin{equation}
    \bm q^j = \begin{bmatrix}
        \cos\frac{\theta^j}{2}\cos\frac{\alpha^j}{2}, & \cos\frac{\theta^j}{2}\cos\frac{\alpha^j}{2}, & 
        \cos\frac{\theta^j}{2}\cos\frac{\alpha^j}{2}, &
        \cos\frac{\theta^j}{2}\cos\frac{\alpha^j}{2} 
    \end{bmatrix}.
\end{equation}
\label{app:arm_model}
The generic kinematic model in \cref{eq:Tquat} can be expressed as
\begin{equation}
    \bm{g}_e(\bm\mu_i^{j}, \bm\mu_e^{j-1}) = \begin{bmatrix}
        \bm t^{j-1} + \bm h(\bm q^{j-1}\cdot[0\ \bm t^j]\cdot\bm q^{j-1*}) \\
        \bm q^{j-1}\cdot \bm q^{j}
    \end{bmatrix}, = \begin{bmatrix}
        x^{j-1} + x_{tf} \\
        y^{j-1} + y_{tf} \\
        z^{j-1} + z_{tf} \\
        q_{w,\ tf} \\
        q_{x,\ tf} \\
        q_{y,\ tf} \\
        q_{z,\ tf} 
    \end{bmatrix},
\end{equation}
where $x_{tf}, y_{tf}, z_{tf}$ and $q_{*,\ tf}$ are the transformed translations and rotation. Computing the Hamilton products yields the following expressions for the transformed positions
\begin{eqnarray*}
x_{tf} &=& {q_w^{j-1}}^2 l^j\cos\theta^j + {q_x^{j-1}}^2 l^j\cos\theta^j - {q_y^{j-1}}^2 l^j\cos\theta^j - {q_z^{j-1}}^2 l^j\cos\theta^j \\
&& + 2q_x^{j-1}q_y^{j-1} l^j\sin\theta^j + 2q_x^{j-1}q_z^{j-1} d^j + 2q_w^{j-1}q_y^{j-1} d^j - 2q_w^{j-1}q_z^{j-1} l^j\sin\theta^j, \\
y_{tf} &=& {q_w^{j-1}}^2 l^j\sin\theta^j - {q_x^{j-1}}^2 l^j\sin\theta^j + {q_y^{j-1}}^2 l^j\sin\theta^j - {q_z^{j-1}}^2 l^j\sin\theta^j \\
&& + 2q_x^{j-1}q_y^{j-1} l^j\cos\theta^j + 2q_y^{j-1}q_z^{j-1} d^j - 2q_w^{j-1}q_x^{j-1} d^j + 2q_w^{j-1}q_z^{j-1} l^j\cos\theta^j, \\
z_{tf} &=& {q_w^{j-1}}^2 d^j - {q_x^{j-1}}^2 d^j - {q_y^{j-1}}^2 d^j + {q_z^{j-1}}^2 d^j \\
&& + 2q_x^{j-1}q_z^{j-1} l^j\cos\theta^j + 2q_y^{j-1}q_z^{j-1} l^j\sin\theta^j - 2q_w^{j-1}q_y^{j-1} l^j\cos\theta^j + 2q_w^{j-1}q_x^{j-1} l^j\sin\theta^j,
\end{eqnarray*}
and orientation:
\begin{eqnarray*}
q_{w,\ tf} &=& q_w^{j-1} \cos\frac{\theta^j}{2}\cos\frac{\alpha^j}{2}
- q_x^{j-1} \cos\frac{\theta^j}{2}\sin\frac{\alpha^j}{2}
- q_y^{j-1} \sin\frac{\theta^j}{2}\sin\frac{\alpha^j}{2} 
- q_z^{j-1} \sin\frac{\theta^j}{2}\cos\frac{\alpha^j}{2}, \\
q_{x,\ tf} &=& q_w^{j-1} \cos\frac{\theta^j}{2}\sin\frac{\alpha^j}{2}
+ q_x^{j-1} \cos\frac{\theta^j}{2}\cos\frac{\alpha^j}{2}
+ q_y^{j-1} \sin\frac{\theta^j}{2}\cos\frac{\alpha^j}{2}
- q_z^{j-1} \sin\frac{\theta^j}{2}\sin\frac{\alpha^j}{2}, \\
q_{y,\ tf} &=& q_w^{j-1} \sin\frac{\theta^j}{2}\sin\frac{\alpha^j}{2} 
- q_x^{j-1} \sin\frac{\theta^j}{2}\cos\frac{\alpha^j}{2}
+ q_y^{j-1} \cos\frac{\theta^j}{2}\cos\frac{\alpha^j}{2}
+ q_z^{j-1} \cos\frac{\theta^j}{2}\sin\frac{\alpha^j}{2}, \\
q_{z,\ tf} &=& q_w^{j-1} \sin\frac{\theta^j}{2}\cos\frac{\alpha^j}{2}
+ q_x^{j-1} \sin\frac{\theta^j}{2}\sin\frac{\alpha^j}{2} 
- q_y^{j-1} \cos\frac{\theta^j}{2}\sin\frac{\alpha^j}{2}
+ q_z^{j-1} \cos\frac{\theta^j}{2}\cos\frac{\alpha^j}{2}.
\end{eqnarray*}

\subsection{Collision avoidance in HAIF}
\label{app:collision_avoidance}
A repulsive state $\bm \mu^!$ can be imposed on intrinsic beliefs, to realize joint limit avoidance, or extrinsic beliefs for collision avoidance with the environment. We define joint limit avoidance as:
\begin{equation}
    \bm{f}_{r,\theta}(\bm{\mu}) =
    \begin{cases}
    0, & \text{if } ||\bm e_\theta|| > \gamma_\theta \\
    k_{r,\theta}\zeta(1/\gamma_\theta - 1/||\bm e_\theta||), & \text{otherwise}
\end{cases},
\end{equation}
where $\bm e_\theta = \bm\mu_\theta^! - \bm\mu_\theta$, $\bm \mu_\theta$ is the slice of beliefs about joint angles, $ \bm\mu_\theta^!$ are the joint limits, and $\gamma_\theta$ is a chosen threshold. The variable $\zeta\in\{-1,\ 1\}$ is negative for lower limits and positive for upper limits. The collision avoidance strategy is instead the same as \cite{priorelli_deep_2023}:
\begin{equation}
    \bm{f}_{r,obst}(\bm{\mu}) =
    \begin{cases}
    0, & \text{if } ||\bm e_{obst}|| > \gamma_{obst} \\
    k_{r,obst}(1/\gamma_{obst} - 1/||\bm e_{obst}||)\bm e_{obst}/||\bm e_{obst}||^3, & \text{otherwise}
\end{cases},
\end{equation}
where $\bm e_{obst} = \bm\mu_{pos}^! - \bm\mu_{pos}$, $\bm \mu_{pos}$ is the slice of beliefs about link positions, and $ \bm\mu_{pos}^!$ is the position of an obstacle given by the VBGS module. Goal attractors and repulsive forces for joint limits and collision avoidance are then summed together to form the dynamics function of a single level. This allows one to achieve behaviors such as reaching a target while avoiding an obstacle. Parameters are manually chosen to achieve sufficient performance in the test cases, but could be automatically optimized. 

\subsection{Whole-body Skills}
\label{app:skills}
The routines for the different skills for the mobile manipulator are defined as follows:
\begin{itemize}
    \item \texttt{Pick}: The robot unfolds its arm (joint goal), moves to a pre-grasp position above the target object (end-effector + joint goal), and then proceeds to the grasp pose to perform the grasp once close enough (end-effector goal). After grasping, it retreats to the post-grasp pose (end-effector goal) and folds the arm back into a compact configuration (joint goal) (see Fig.~\ref{fig:pick_skill}). 
    \begin{figure}[h!]
        \centering
        \includegraphics[width=\linewidth]{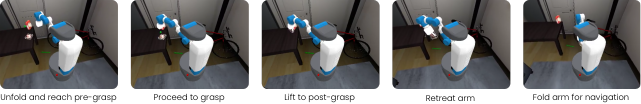}
        \caption{Evolution of the \texttt{Pick} skill over time.}
        \label{fig:pick_skill}
    \end{figure}
    \item \texttt{Place}: It mirrors the \texttt{Pick} sequence, but targets a specified place location.
    \item \texttt{PickFromDrawer}: The end-effector is moved in front of the drawer hinge and grasps the handle once close enough (end-effector + joint goal). Then, the robot executes a linear backward trajectory to pull the drawer open (end-effector goal). The object is picked as in \texttt{Pick}. Finally, the robot end-effector is placed in front of the handle again (end-effector goal), and the drawer is pushed close following a linear trajectory (end-effector goal) (see Fig.~\ref{fig:pick_drawer_skill}). 
    \begin{figure}[h!]
        \centering
        \includegraphics[width=\linewidth]{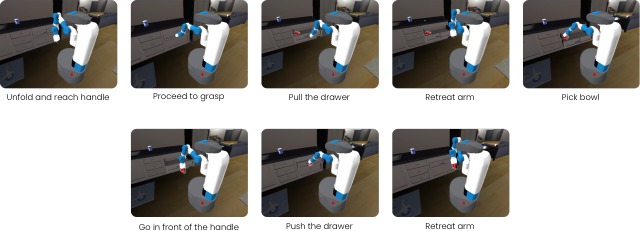}
        \caption{Evolution of the \texttt{PickFromDrawer} skill over time.}
        \label{fig:pick_drawer_skill}
    \end{figure}
    \item \texttt{PickFromFridge}: The robot unfolds its arm (joint goal), moves in front of the fridge handle, and grasps it once close enough (end-effector goal). It then follows a circular trajectory to partially open the door (end-effector goal). After that, the arm retreats (joint goal), and finally, the arm starts a linear trajectory from behind the half-opened door to push it to fully open (end-effector goal). The object is picked as in \texttt{Pick}, and then
   the robot first moves to the left of the fridge door (base + joint goal), and after it follows a linear trajectory to push the door closed (end-effector goal) (see Fig.~\ref{fig:pick_fridge_skill}).
   \begin{figure}[h!]
        \centering
        \includegraphics[width=\linewidth]{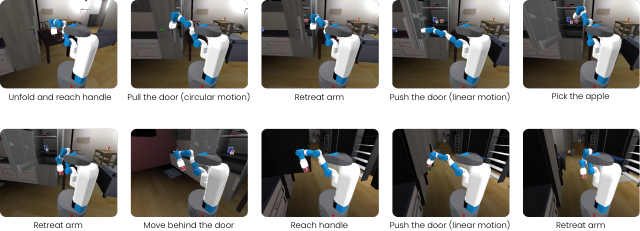}
        \caption{Evolution of the \texttt{PickFromFridge} skill over time.}
        \label{fig:pick_fridge_skill}
    \end{figure}
    \item \texttt{Move}: The \texttt{NavModel} computes a global path towards a final goal and orientation, and provides the move skill with the current active subgoal $(x^*,\ y^*)$, along with the final desired position and orientation. At each step, the heading $\theta^*$ toward the subgoal is computed. A predefined joint configuration (joint goal) for the arm is set to avoid collisions. The skill terminates when the robot is within a threshold distance of the target pose. See Fig.~\ref{fig:move_skill}) for an example. The reach threshold for the position is kept at 0.8m while the one for orientation to 0.3rad. These are particularly loose since we rely on whole-body manipulation skills and are not required to precisely position the base before executing them.
    \begin{figure}[h!]
        \centering
        \includegraphics[width=0.4\linewidth]{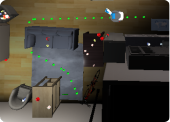}
        \caption{Top view of the \texttt{Move} skill where the robot moves through subgoals following the global path.}
        \label{fig:move_skill}
    \end{figure}
\end{itemize}

\subsection{Example evolution of a probabilistic map}
\begin{figure}[ht!]
    \centering
    \includegraphics[width=.9\linewidth]{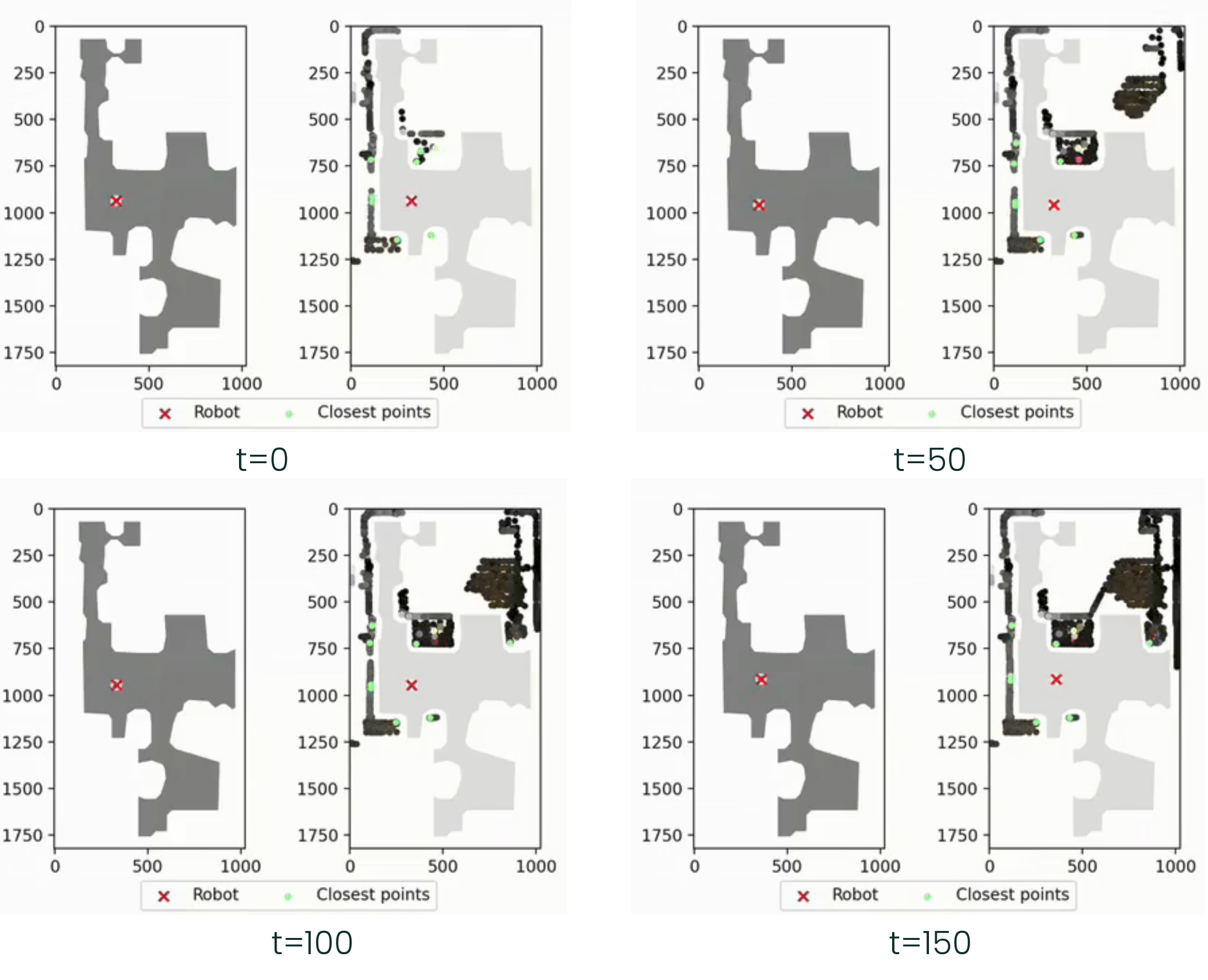}
    \caption{An example of the probabilistic map evolution with VBGS in one Habitat apartment. The left side of each panel shows the location of the robot on the ground truth floor plan. The right side overlays the Gaussian components over the obstacles projected onto the floor.}
    \label{fig:probmap-ex}
\end{figure}

In Fig.~\ref{fig:probmap-ex} we present an example of how a probabilistic map can evolve through time using VBGS. 

\end{document}